%% file: root.tex
\documentclass[letterpaper, 10pt, conference]{ieeeconf}
\IEEEoverridecommandlockouts
\usepackage{amsmath,amssymb,amsfonts}
\usepackage{algorithmic}
\usepackage{graphicx}
\usepackage{subcaption}
\usepackage{textcomp}
\usepackage{xcolor}
\usepackage[hidelinks]{hyperref}    % Clickable citations, figures
\usepackage[nameinlink]{cleveref}   % Full names for figure refs
\crefname{figure}{Fig.}{figures}
\crefname{section}{Section}{sections}
\usepackage[
    backend=biber,
    style=ieee,
    sorting=none
]{biblatex}
\title{\LARGE \bf
Automated Vehicle Platform with Connected Driving Capabilities
}

\author{
    Oskars Teikmanis, \textit{Student Member, IEEE},
    Aleksandrs Levinskis, \textit{Student Member, IEEE},\\ 
    Andris Ivars Mackus,
    Artis Rušiņš,
    Amr Elkenawy,
    Marta Tropa,
    Modris Greitans
\thanks{This work is supported by the “5th Generation connected and automated mobility cross-border EU trials” (5G-ROUTES) project, which has received funding from the European Union's Horizon 2020 Research and Innovation programme under grant agreement No. 951867. The authors are with the Institute of Electronics and Computer Science (EDI), Riga, Latvia. Contact: \texttt{oskars.teikmanis@edi.lv}.}
}

\begin{document}

\maketitle

\begin{abstract}
 Augmenting automated vehicles to wirelessly detect and respond to external events before they are detectable by onboard sensors is crucial for developing context-aware driving strategies. To this end, we present an automated vehicle platform, designed with connectivity, ease of use and modularity in mind, both in hardware and software. It is based on the Kia Soul EV with a modified version of the Open-Source Car Control (OSCC) drive-by-wire module, uses the open-source Robot Operating System (ROS and ROS 2) in its software architecture, and provides a straightforward solution for transitioning from simulations to real-world tests. We demonstrate the effectiveness of the platform through a synchronised driving test, where sensor data is exchanged wirelessly, and a model-predictive controller is used to actuate the automated vehicle.
\end{abstract}

% \begin{IEEEkeywords}
%     Connected automated mobility, research platform, architecture, drive-by-wire, ROS, ROS 2, V2X, 5G.
% \end{IEEEkeywords}

\input{sections/introduction.tex}

\input{sections/related_work.tex}

\input{sections/the_vehicle.tex}

\input{sections/software.tex}

\input{sections/evaluation.tex}

\input{sections/conclusions.tex}

\printbibliography

\end{document}

%% file: sections/introduction.tex
\section{Introduction}

The earliest platforms for ground-up automated driving research have been around since the nineteen-eighties, with a notable example by Dickmanns \textit{et al.} \cite{dickmanns1987autonomous}. Some of the most prominent early attempts at "hacking" a passenger vehicle to make it autonomous have happened in the mid-nineteen-nineties, with pioneering work on vision-based, road-capable vehicles across multiple continents \cite{dickmanns1994seeing, jochem1995pans, lee1996development}. In more recent years, it has become common practice for research institutions to either build their own specialised vehicle \cite{bunte2014central, rassolkin2018development, feher2020highly, arango2020drive} or modify an existing one for research work \cite{munir2018autonomous, scholliers2018development, novickis2020functional, reke2020self, certad2023jku, camara2022openpodcar, brown2022development, mehr2022x}, since the field of autonomous driving has repeatedly proven itself to be highly impactful and complex. However, it has also become increasingly clear that city-wide autonomous mobility is unlikely to be a product of independent agents working out their planning individually. For truly augmented mobility in highly developed areas, the communication between vehicles and other infrastructure is expected to play a meaningful role, and this is reflected in a lot of recent research, whether it concerns platforms for connected vehicle development and validation \cite{scholliers2018development, gand2020lightweight, mehr2022x, avedisov2018experimental} or higher-level city-planning \cite{ismagilova2019smart}.

With these considerations in mind, we present in this paper a platform for connected autonomous driving, explaining its design philosophies, both on a hardware and software level. We also highlight some of its more distinctive features, including an accent on connected driving, aiming for robustness against crossing country borders.

The remainder of this document is organised as follows: Section \ref{sec:related_work} discusses the latest developments in the field; Section \ref{sec:vehicle} introduces the vehicle and its hardware architecture; Section \ref{sec:software} describes the software architecture, including aspects related to connected driving. Subsequently, Section \ref{sec:evaluation} demonstrates the testing of our vehicle in both virtual and real-world environments. Lastly, Section \ref{sec:conclusions} offers concluding remarks on the accomplishments and highlights ongoing and future research directions.

%% file: sections/related_work.tex
\section{Related Work} \label{sec:related_work}

With the advent of electric vehicles, especially in combination with affordable sensors and open-source drive-by-wire (DbW) systems, it has become cheaper and much more convenient to retrofit a vehicle with the fundamentals for autonomous driving research. 

Some researchers build their vehicles from the ground up, and equip them with unique driving functionalities, like the over-actuated \textit{RoboMobil} by Bünte \textit{et al.} whose main design philosophy is to enable advanced vehicle dynamics research \cite{bunte2014central}. However, it is not uncommon for such platforms to be built with the goal of acquiring deeper experience in the field (often for educational purposes), and being in control of nearly every technical process \cite{feher2020highly, rassolkin2018development, arango2020drive}. Here we would like to highlight the work by Arango \textit{et al.} \cite{arango2020drive}, as it presents an extensive overview of their compact electric vehicle platform, which has been made open-source. It includes details on the used hardware, system and software architecture, DbW system, controllers and driving performance.

Still, such custom platforms are unlikely to represent the driving behaviour that can be expected from road-legal vehicles, and it is evident that the path chosen by many is to use an existing vehicle as the base for the platform \cite{ulmer1994vita, jochem1995pans, lee1996development, scholliers2018development, reke2020self, munir2018autonomous, certad2023jku, camara2022openpodcar, brown2022development, mehr2022x}. Common design choices include simplicity, modularity and affordability. Some put an additional focus on special architectural aspects, such as Robot Operating System (ROS) integration \cite{reke2020self, munir2018autonomous, certad2023jku, camara2022openpodcar, brown2022development, mehr2022x}. In most cases, the first version of ROS (hereinafter - ROS 1) appears to be the chosen version due to the availability of libraries for sensor integration and other interfaces, despite its lack of real-time system compatibility. Also, since road safety is largely ensured via controlled test environments and manual override solutions, the limitations of ROS 1 are mostly compensated for, at least in a research environment. However, there are examples of at least partial ROS 2 usage in autonomous vehicle platforms \cite{reke2020self, certad2023jku, mehr2022x}. The potential of the updated version has been shown via timing analysis experiments \cite{maruyama2016exploring, reke2020self, kronauer2021latency}. While the usage of the standardised Data-Distribution Service \cite{pardo2003omg} makes ROS 2 clearly superior to ROS 1 in terms of timing behaviour, it still has an overhead that depends on different factors such as the used hardware, transmitted payload size and publishing frequency.

A common trend in autonomous vehicle development has been the increasing focus on cooperative, connected and autonomous mobility (CCAM), which includes functionalities that let vehicles communicate among themselves (V2V) or with any external infrastructure element (V2X). This has also been reflected in some of the existing platform reports \cite{certad2023jku, mehr2022x, scholliers2018development}. Scholliers \textit{et al.} \cite{scholliers2018development} use a connected vehicle equipped with a 5G-compatible on-board-unit (OBU) to implement a collision-avoidance system that lets traffic participants share perception data in real-time on a shared map. The numerous finished and ongoing \textit{Horizon 2020} projects centred around 5G applications in mobility and logistics (5G-CARMEN \cite{5g-carmen}, 5G-MOBIX \cite{5g-mobix}, 5G-ROUTES \cite{5g-routes}, 5G-BLUEPRINT \cite{5g-blueprint}, to name a few) give a strong indication about the recognised value of connected systems in the future of mobility applications in the European Union.

%% file: sections/the_vehicle.tex
\section{The Vehicle} \label{sec:vehicle}

When designing our research vehicle platform, we defined requirements that fit its envisioned usage. They can be summarised as follows:
\begin{enumerate}
    \item Usability: the platform must be simple to use, both in terms of adding new hardware, developing new software and network functions, and executing them.
    \item Scalability: hardware and software additions to fulfil new research demands must be easy and logical.
    \item Transparency: the entire hardware and software design of the platform must be comprehensive, reproducible.
\end{enumerate}

\begin{figure}[htbp]
    \centerline{\includegraphics[width=\linewidth]{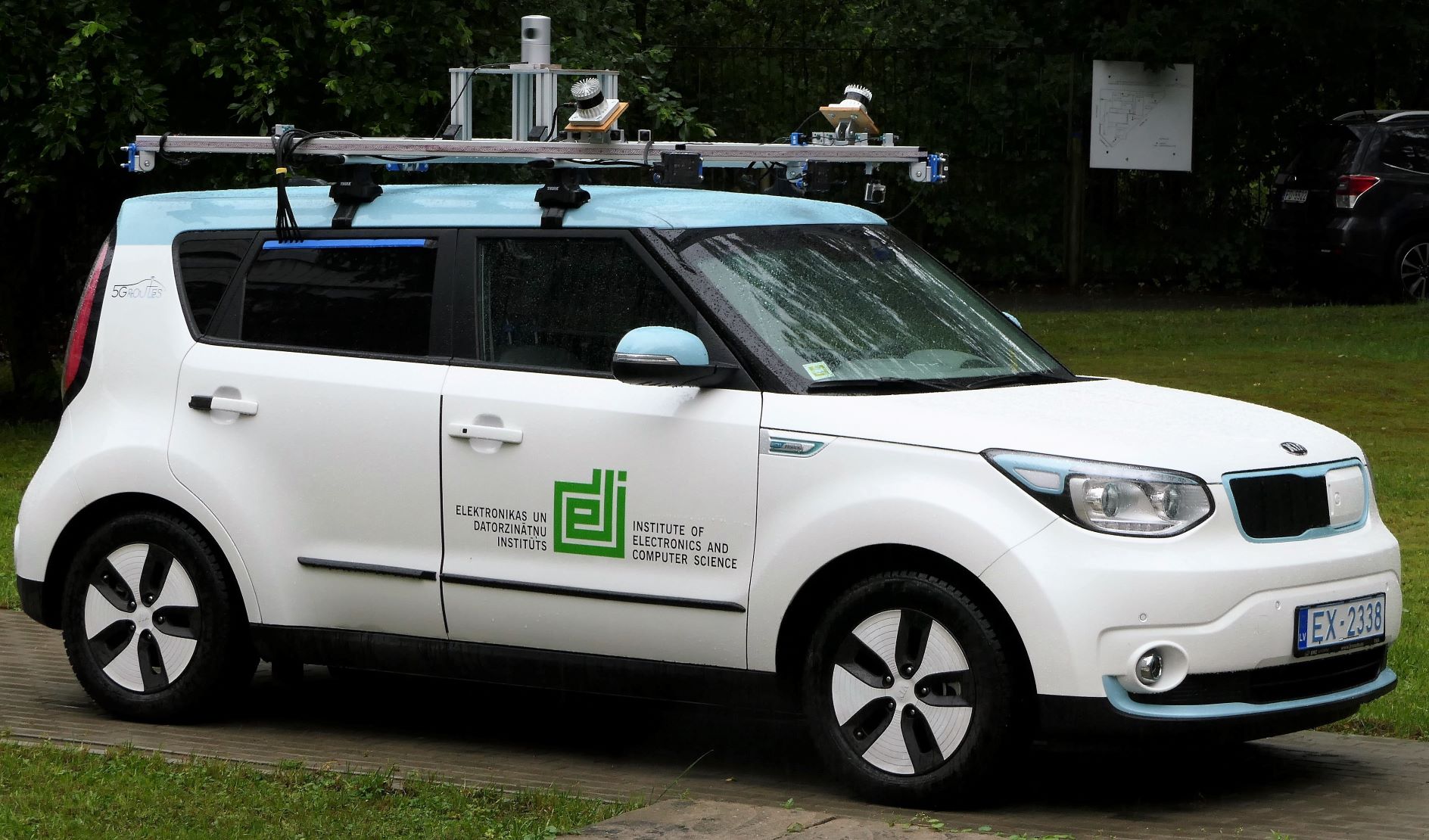}}
    \caption{The \textit{White Kia}. Its primary set of sensors consists of a lidar (on the roof platform), a radar (behind the nose cover), a stereo camera (behind the windshield) and a GPS-RTK receiver (near the back of the vehicle's roof). The roof platform also provides space for additional sensors such as surround view cameras, radars and side-mounted lidars, which are also featured in the photograph.}
    \label{fig:white-kia}
\end{figure}

The Kia Soul is a fitting candidate for these considerations, as it has a convenient form-factor for hardware augmentations, and is compatible with the Open-Source Car Control (OSCC) system \cite{oscc-github}. The model we selected for modifications is a 2017 Kia Soul EV (\cref{fig:white-kia}), hereinafter - the \textit{White Kia}. It serves as a platform for various research areas, including driving algorithms, perception and, most recently, connected mobility, across multiple projects \cite{ai4csm, 5g-routes, augmented-ccam}. For connected driving capabilities, we also utilise a sister vehicle of the same model, the \textit{Blue Kia}. The initial stages of defining and implementing the functional architecture of the \textit{Blue Kia} are outlined in \cite{novickis2020functional}. This section is dedicated to the design considerations and hardware of the \textit{White Kia}.

\subsection{Perception Hardware}
\begin{figure}[htbp]
    \centerline{\includegraphics[width=\linewidth]{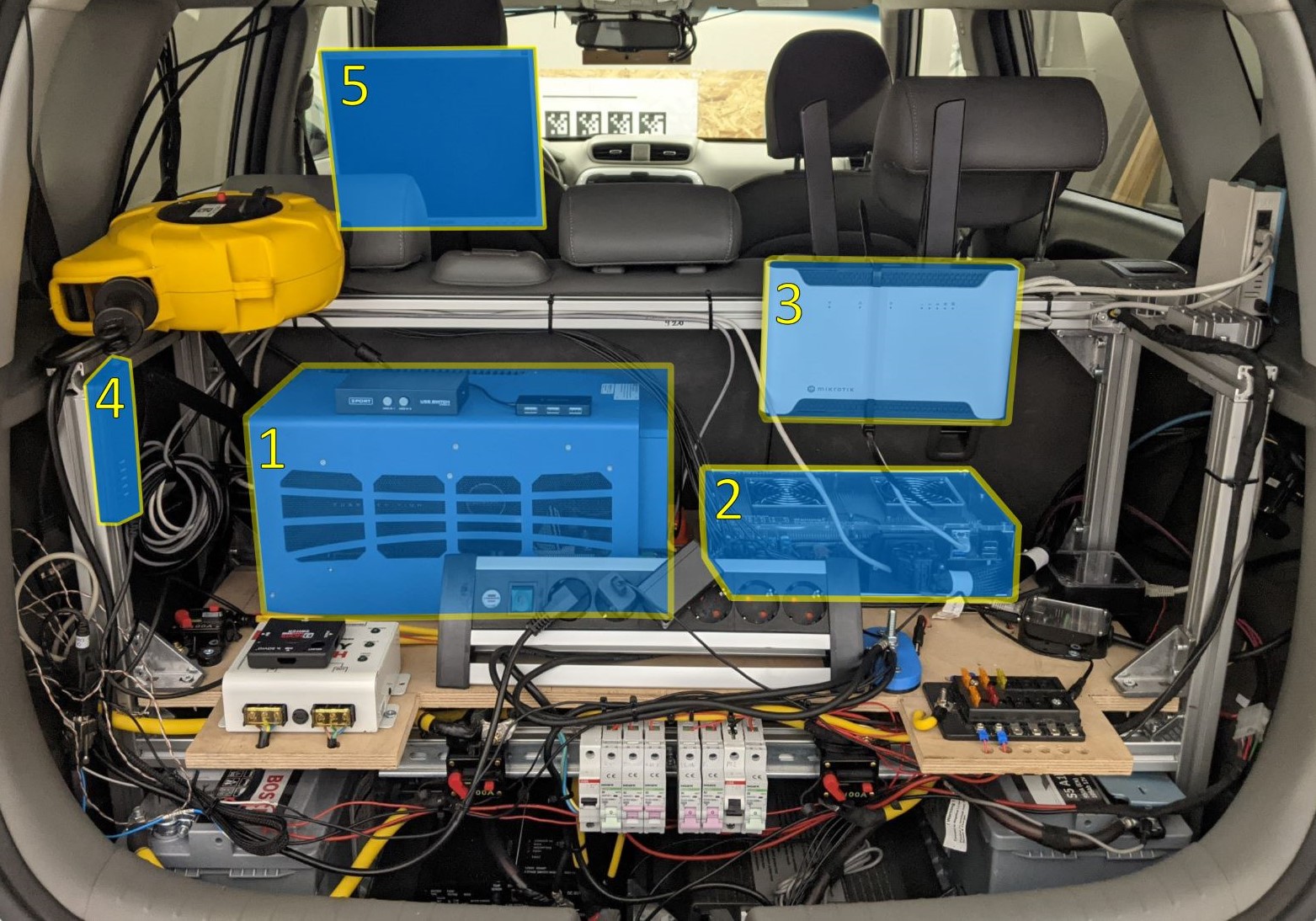}}
    \caption{Boot setup. Marked in blue are the computing systems and peripherals. The rugged Nuvo-8108GC-XL (1) serves as the main computer of the vehicle, and communicates with most sensors, except for the GMSL cameras which are connected to the Nvidia Drive PX2 (2). For network connectivity we use the Mikrotik Chateau 5G modem (3). CAN-based components use the Kvaser USBLight adapter (4). The monitor (5) is connected to both on-board computers. The bottom area contains batteries and power electronics, including an inverter, mains priority switch and two 12V/100Ah batteries.}
    \label{fig:boot-setup}
\end{figure}

\begin{figure*}[htbp]
    \centerline{\includegraphics[width=\linewidth]{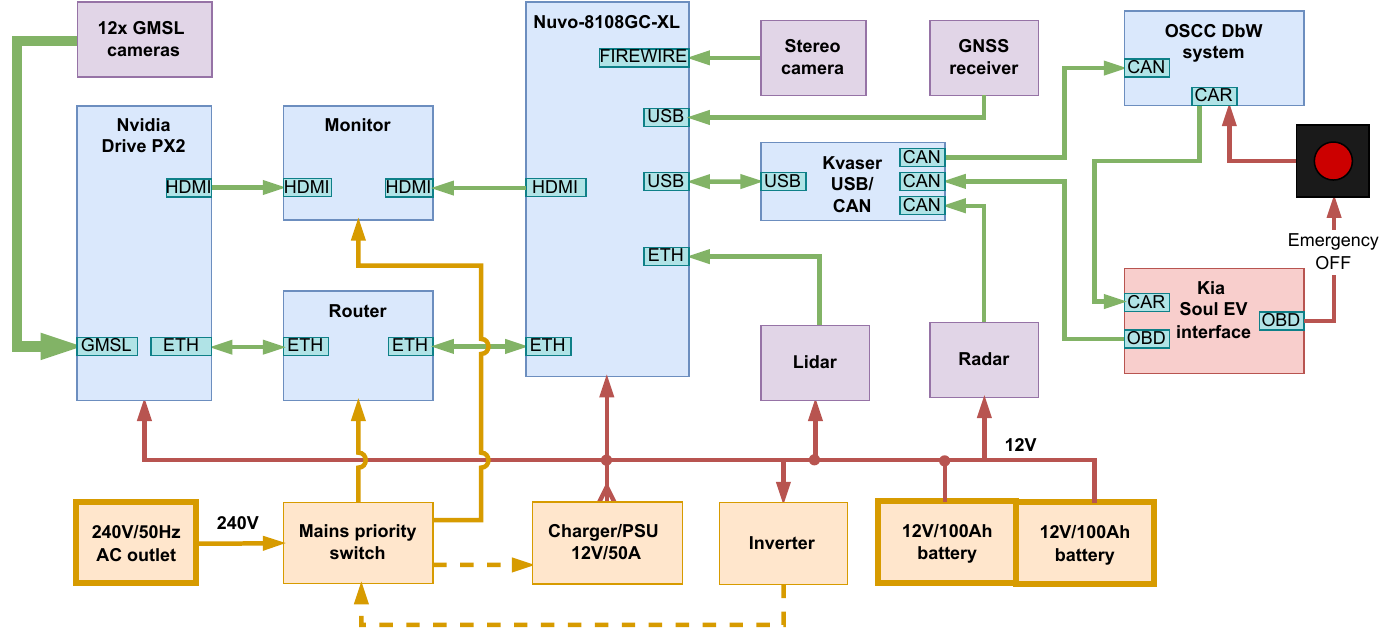}}
    \caption{Hardware schematic. The computing systems and peripherals are marked in blue, while the sensors are in purple and the power electronic components are in yellow. The red, yellow and green wires represent the 12V, 240V supply and data flow respectively. The vehicle interface is marked in red. It powers the DbW system through the vehicle’s OBD2 port, and transmits internal sensor data (IMU, encoder etc.) directly to the main computer via CAN-to-USB. All other components are either powered by the supplementary 12V batteries or a wall outlet. If the system is connected to a 240V outlet, the charger is active, otherwise the inverter powers the on-board
electronics.}
    \label{fig:hw-schematic}
\end{figure*}

The \textit{White Kia} has been modified in three major areas: the roof, boot space and driver area. The roof contains most of the sensors, all located on an aluminium frame, which is attached to a Thule roof-rack (featured in \cref{fig:white-kia}). The boot of the vehicle contains most of the computation units and power electronics (\cref{fig:boot-setup}). A detailed hardware schematic is shown in \cref{fig:hw-schematic}. Some of the main sensors we use are the following:

\begin{itemize}
    \item \textbf{ARS 408-21} long-range radar for tracking objects longitudinally. This model operates at 77 GHz with a bandwidth of 1 GHz. It is commonly used in automotive applications and automated vehicle platforms \cite{feher2020highly, mehr2022x}.
    \item \textbf{Velodyne HDL-32E} lidar for localisation and perception. We also used it as a basis for building high definition (HD) maps of our main test environments.
    \item \textbf{Bumblebee XB3} front-facing stereo camera for depth vision and perception. It features three 1.3 MP sensors with two baselines (12 cm and 24 cm).
    \item \textbf{Emlid Reach M+} single band RTK GNSS module for highly accurate localisation, using correction data from RTK base-stations. We use it for improving the accuracy of our larger HD maps via geo-referencing.
\end{itemize}

For safety, the hardware system is equipped with two solutions: (1) a manual override function that returns control to the driver as soon as any user input is given to the pedals or steering wheel, and (2) an emergency switch near the driver's seat, that cuts the power to the DbW system when pressed. It is intended to be used if the manual override fails to engage.

\subsection{\label{sec:dbw-system} Drive-by-Wire System}
\begin{figure*}[htbp]
  \centering
  \begin{subcaptionblock}{0.495\linewidth}
    \includegraphics[width=0.95\linewidth]{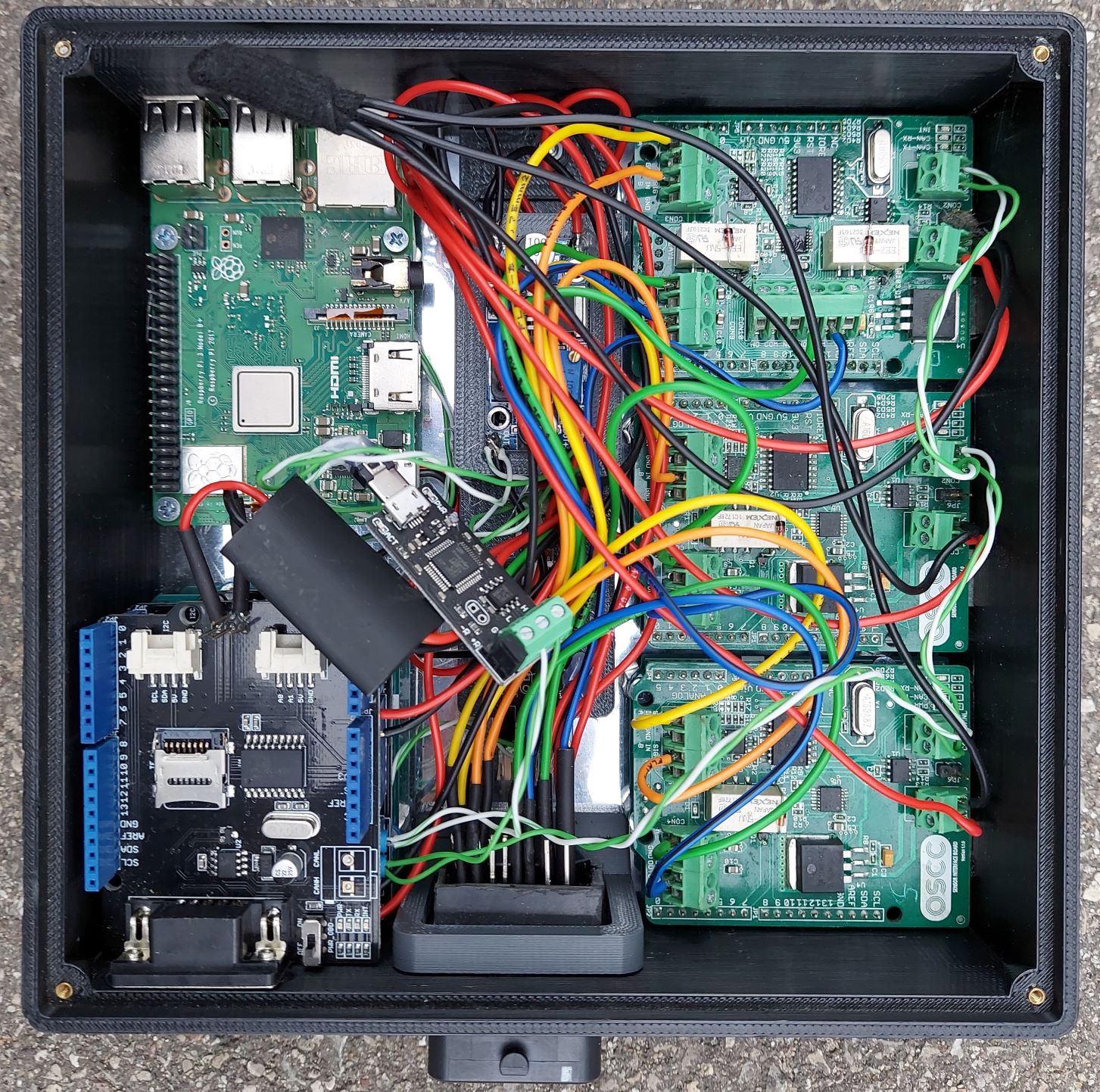}
  \end{subcaptionblock}
  \begin{subcaptionblock}{0.495\linewidth}
    \includegraphics[width=1\linewidth]{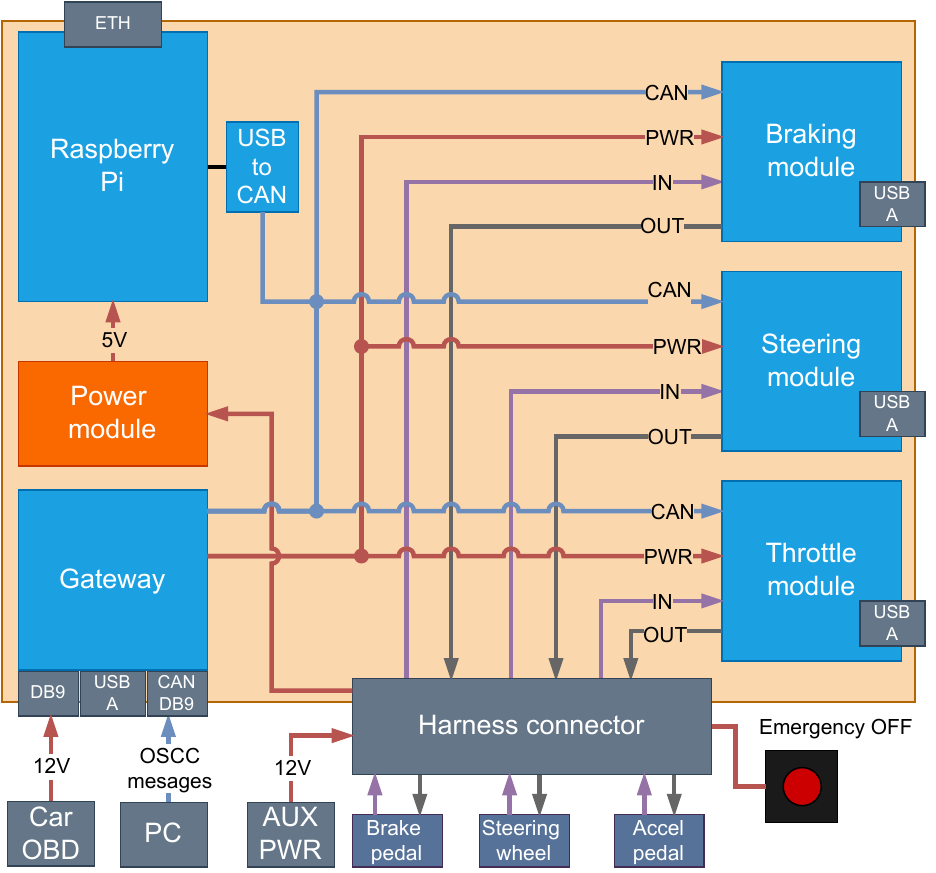}
  \end{subcaptionblock}
  \caption{Modified OSCC hardware unit (left) and its corresponding architecture schematic (right). The original design has 4 modules: a gateway for passing OBD messages to the OSCC CAN bus, throttle module, steering module and braking module. In addition to these modules, we have added a Raspberry Pi as an auxiliary processing unit, connected to the OSCC CAN bus via USB-CAN converter, enabling our version of the OSCC system to be used without the need for an external CAN adapter.}
  \label{fig:oscc}
\end{figure*}

We opted for OSCC as our preferred system for implementing automated throttle, brake, and steering actuation, because it is both open-source and compatible with the Kia Soul EV. OSCC uses a form of signal spoofing, and acts as a intermediary between the vehicle and its electronic control unit (ECU). Next to its core modules (gateway, steering, throttle and braking), we incorporated our own modifications in the form of a Raspberry Pi, which serves as an auxiliary processing unit, as illustrated in \cref{fig:oscc}.

OSCC connects to the vehicle's OBD port, which draws power from the 12V battery. However, due to the limited amperage of this configuration, it is insufficient to power the Raspberry Pi. To solve this, we added an additional connection to the vehicle's 12V socket, which supplies up to 10A.

%% file: sections/software.tex
\section{Software} \label{sec:software}

The same high-level requirements, such as ease of use, modularity, transparency, and the possibility for communication services that define the hardware setup, were also considered in the software design. To ensure that, we favour loose coupling and open-sourceness of software components, for simple integration and reconfiguration between simulations and real-world tests.

\subsection{Architecture}
\begin{figure*}[htbp]
    \centerline{\includegraphics[width=\linewidth]{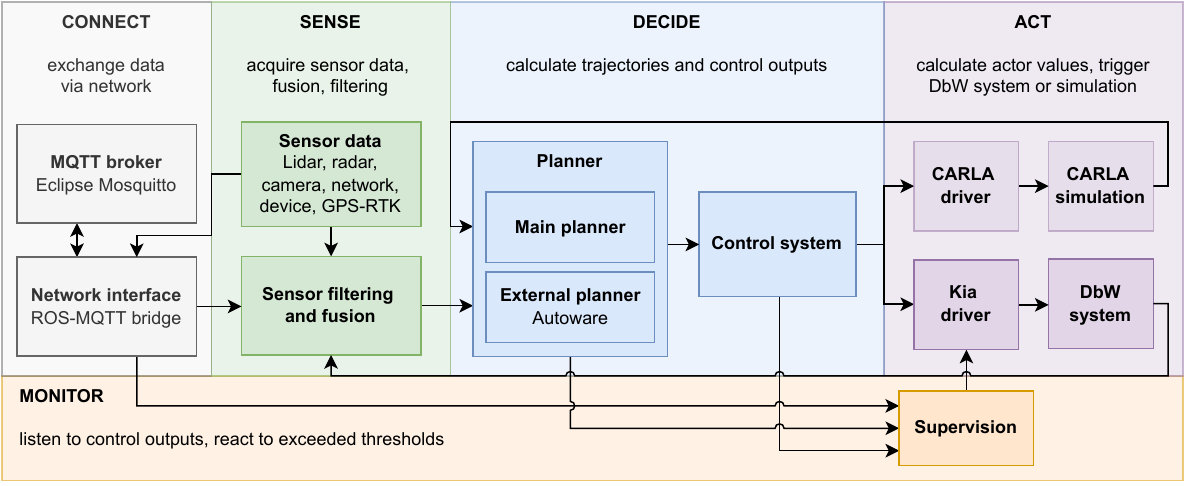}}
    \caption{Software architecture. Our information flow structure follows the \textit{sense-decide-act} paradigm, which is very common in robotics. We also integrated communication infrastructure into the loop, effectively turning our network device into an additional sensor that can obtain any type of data through an MQTT broker and then feed it into perception stack. After calculating control outputs for the next time step, our architecture offers two paths: a virtual driver for simulated vehicles in the CARLA environment, and a Kia driver that produces the signals that are directly fed into the DbW system. In both cases this driver is responsible for translating control signals into low-level actuation signals.}
    \label{fig:sw-arch}
\end{figure*}

Due to the interplay between different sensors and communication interfaces, we chose the blackboard design pattern in our software architecture, and implemented it primarily via ROS~2, using a mix of C++ and Python. This decision was motivated by the fact that there is a strong community developing numerous software packages for this framework, including middleware for network protocols and sensor integration. We also make use of Autoware, an open-source software stack for self-driving vehicles \cite{kato2015open}, initially implemented in ROS~1, and currently in development for ROS~2. In the early stages of conceptualising our software architecture, Autoware was mainly available in ROS~1, so we opted for a hybrid solution that uses the older version alongside ROS~2. A graphical representation of the resulting architecture is shown in \cref{fig:sw-arch}.

Software-side safety solutions include maximum throttle and torque thresholds, and a supervision module that listens to signals from the communication device and control loop, and produces appropriate status signals. The vehicle can then react accordingly if, for example, specific signal flows are interrupted for an extended time. Reactions range from activating backup controllers to disabling the electronic actuation system completely. We also implemented an option to separate the activation of electronic actuation modules. Among other things, this lets us test longitudinal control while letting the driver take care of steering. If this setup is used, the manual override only works when the currently active controls are interacted with by the driver.

\subsection{V2X Communication}
Our automated vehicle platform is developed in a way that it can receive data from external connected devices. To support V2X communication we use MQTT, a lightweight messaging protocol suited for low-bandwidth applications. Data of interest (such as pose, speed, curvature etc.) are packed into cooperative awareness messages (CAMs) \cite{ETSI_CAM} using the ASN1C compiler \cite{asn1c}, converted into MQTT topics (using a ROS-MQTT bridge), and sent to an MQTT broker using \textit{Eclipse Mosquitto} - a lightweight open-source MQTT protocol implementation. The broker, situated externally to the vehicle, can be accessed via internet connectivity supplied by an internet gateway device. This connection enables the vehicle to transmit and receive data from other V2X-enabled devices, as the information on the broker-side is disseminated to all subscribers. The specific device used for internet access bears minimal significance, provided that the MQTT broker and all V2X connected devices are within the same network.

%% file: sections/evaluation.tex
\section{Evaluation} \label{sec:evaluation}

We make use of a multitude of resources to assess and refine our CCAM platform and its hosted algorithms. This section offers a comprehensive overview of the virtual and real environments at our disposal, along with a description of how we have employed them. To demonstrate this, we present a connected driving application that highlights the synergy between the \textit{White Kia's} V2X functionalities and its control system.

\subsection{Synchronised Driving Control} \label{sec:state-sync-mpc}
% MPC controller intro, cost function etc.
% Cost function further explained
One of the connected driving applications developed for our vehicle platform is a model-predictive controller (MPC) that calculates a weighted optimum for speed and acceleration synchronisation between two vehicles: a reference vehicle and a controlled vehicle that obtains control inputs exclusively via wireless communication. The MPC minimises the following cost function:
\begin{equation}
\begin{aligned}
J_i & = \sum_{k=0}^{H} (C_v(v^{ego}_k - v^{ref})^2_i + C_a(a^{ego}_k - a^{ref})^2_i) \\
  & + C_u||\textbf{u}_i - \textbf{u}_{i-1}||^2,
\label{eq:cost-func}
\end{aligned}
\end{equation}
where $v^{ego}_k$ and $a^{ego}_k$ are the predicted speed and acceleration at discrete time prediction step $k$ respectively, and $v^{ref}$ and $a^{ref}$ are the speed and acceleration of the reference vehicle, evaluated at controller sample step $i$. $v^{ref}$ and $a^{ref}$ are obtained wirelessly from the reference vehicle. $\textbf{u}_t$ and $\textbf{u}_{t-1}$ are the control vectors at the current and previous controller steps respectively, and are used for avoiding large jumps in control values. Their size is $\mathbb{R}^{H\times1}$, and $a^{ego}_k = \textbf{u}_t[0]$. Note that $k$ is used by the MPC to iterate through prediction time steps, while $t$ is a time step of the controller. $H$ is the prediction horizon of the MPC. The parameters $C_v$, $C_a$ and $C_u$ are used for setting the relative weight of each term.

For the prediction step of the controller, we use a kinematic model that relates the vehicle's acceleration with the relative speed to the reference vehicle:

\begin{equation}
\begin{aligned}
\Delta v_{k+1} = v^{ego}_k - v^{ref} + (a^{ego}_k - a^{ref}) T_s,
\end{aligned}
\end{equation}
where $T_s$ is the prediction time step of the MPC. These calculations are made for the entire prediction horizon, and fed into \cref{eq:cost-func} along with measured values, to produce an array of control outputs, the first of which is used as the immediate signal that drives the vehicle's actuators. It is worth noting that a model-predictive controller can also be implemented with a dynamic model that takes into account physical properties of the vehicle (such as inertia and motor torque), but such an approach would make it very difficult to generalise the controller for different vehicles. We use a time step $T_s=0.2~s$ and a prediction horizon $H=15$ steps. 

\subsection{Test Environments}

\begin{figure*}[htbp]
    \centerline{\includegraphics[width=\linewidth]{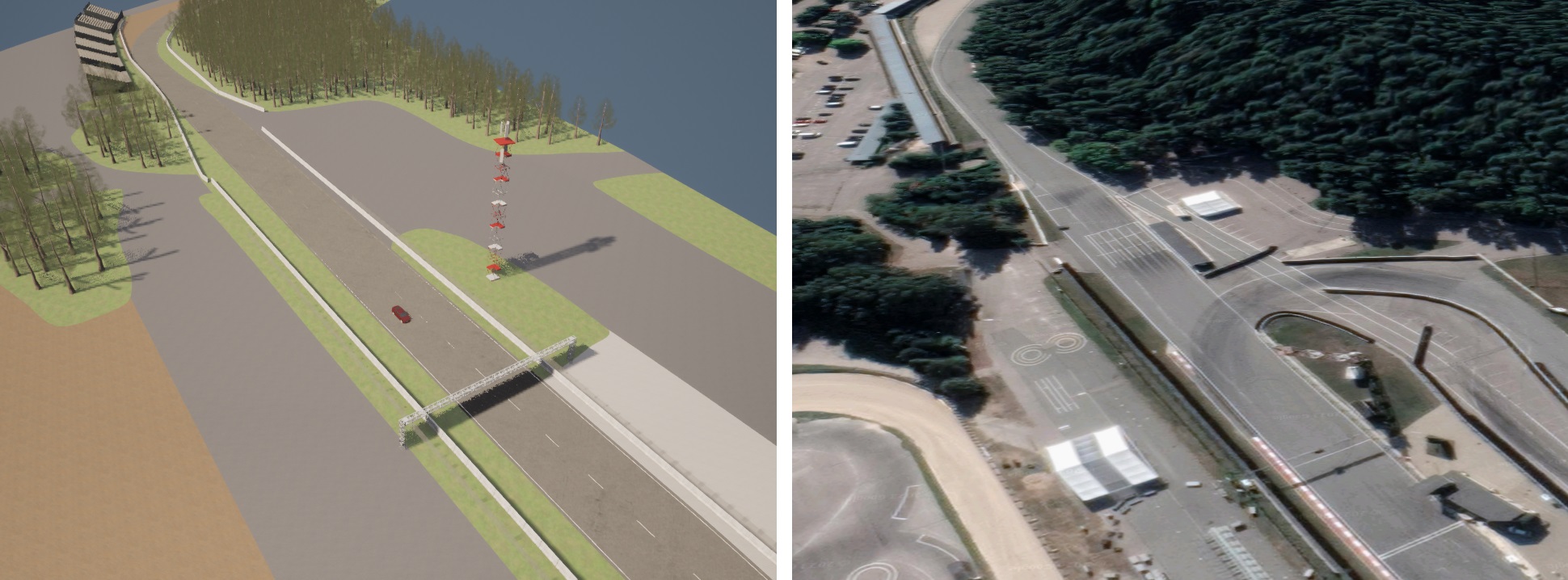}}
    \caption{Test environments. HD map of the \textit{Biķernieki} racetrack (left), bird's-eye view of the real racetrack (right). Source: Google Maps (accessed 15.03.2023).}
    \label{fig:test-envs}
\end{figure*}

\textbf{Simulated racetrack in CARLA}. Prior to implementing our algorithms on the \textit{White Kia's} on-board computer, we conduct simulated driving tests within the CARLA environment. Depending on the specific application, we either opt for a default map or utilise our custom virtual \textit{Biķernieki} racetrack environment (\cref{fig:test-envs}, left). This virtual environment is constructed using the 3D map-making tool \textit{RoadRunner} \cite{roadrunner}, with a lidar-based map of the racetrack serving as its template.

\textbf{\textit{Biķernieki} racetrack in Riga}. The main loop of the racetrack features a wide finish straight and several smaller roads (\cref{fig:test-envs}, right), enabling safe trials with multiple vehicles. A distinctive characteristic of the racetrack is the presence of two cellular base stations: one providing broadband services from a local mobile network operator (LMT), and the second from an Estonia-based provider (Telia). This enables the simulation of a country border-crossing zone \cite{5g-routes-report}. This configuration was specifically implemented to test the network handover functionalities of the 5G infrastructure in border-crossing scenarios, eliminating the necessity for physically driving to a country border area. We use our own MQTT broker which is physically located about 2 km from racetrack. %Funding for this setup was partly provided by the H2020 project \textit{5G-ROUTES}.

\subsection{Driving Tests}

To evaluate the effectiveness of our longitudinal controllers in CARLA environments, we select maps that provide sufficient longitudinal space, and populate them with virtual vehicles. We allow the leading vehicle to be either autonomously driven by the simulator or manually controlled using a sine-like throttle curve. The following vehicles then calculate throttle values based on the MPC. Input values such as speed and acceleration values are obtained directly from the simulator.

\begin{figure}[htbp]
    \centerline{\includegraphics[width=\linewidth]{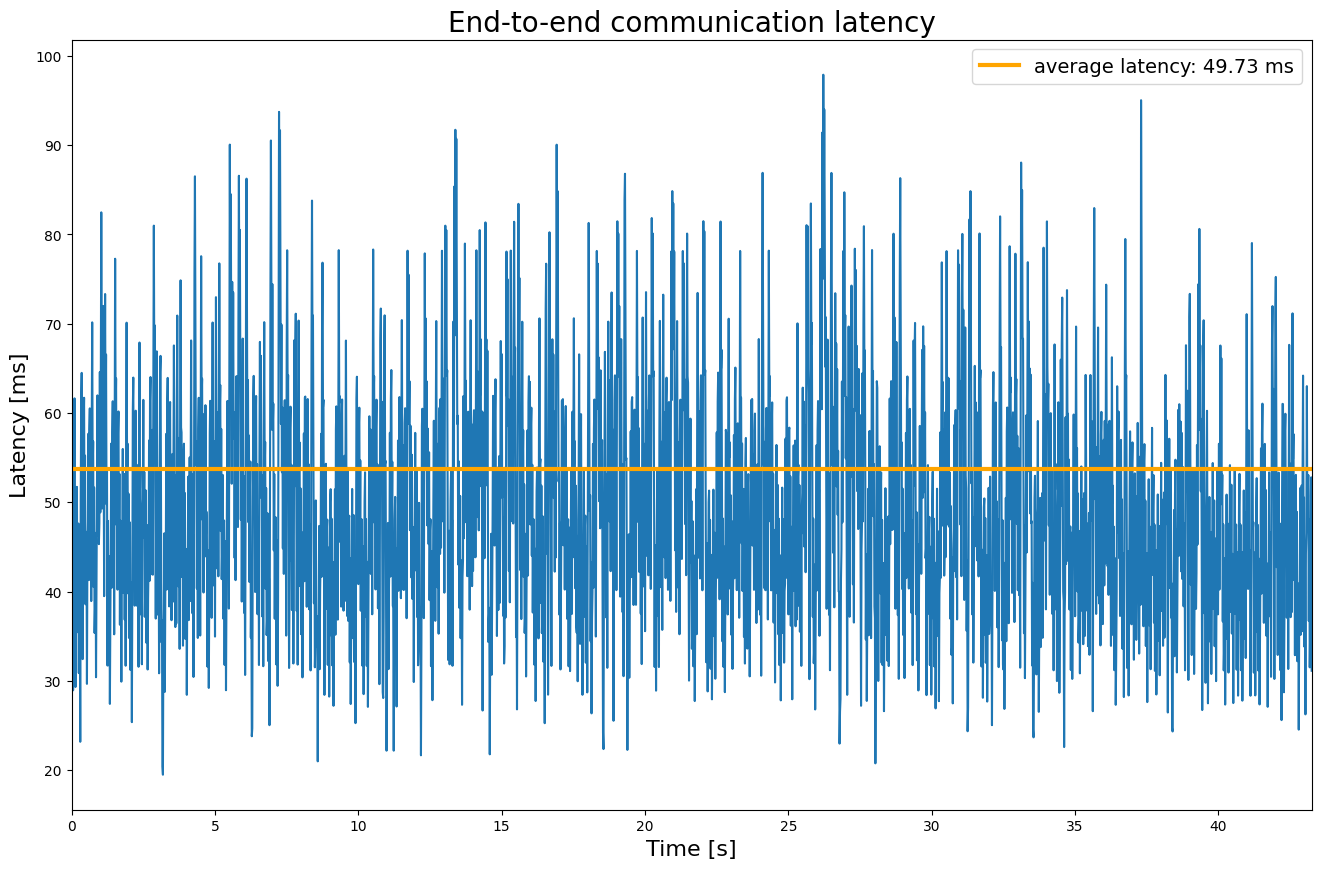}}
    \caption{End-to-end latency measurement of V2X messages exchanged via MQTT and intercepted via ROS.}
    \label{fig:mqtt-plot}
\end{figure}

We conduct analogous tests on the \textit{Biķernieki} racetrack, using our test vehicles - the \textit{White Kia} and \textit{Blue Kia}. Prior to the use of wirelessly exchanged data, we first generate synthetic external sensor data, such as speed and acceleration, to observe driving behaviour. When performing connected driving experiments, end-to-end (E2E) latency of exchanged messages is a critical factor. To measure the E2E latency, we utilise the 5G Non-standalone (NSA) network available on the racetrack. We publish data from the \textit{Blue Kia} to our MQTT broker, subscribe to it from the \textit{White Kia}, and then compare the timestamps encoded in the messages. As shown in \cref{fig:mqtt-plot}, our measurements produce an average latency of 49.73 ms.

\begin{figure*}[htbp]
    \centerline{\includegraphics[width=\linewidth]{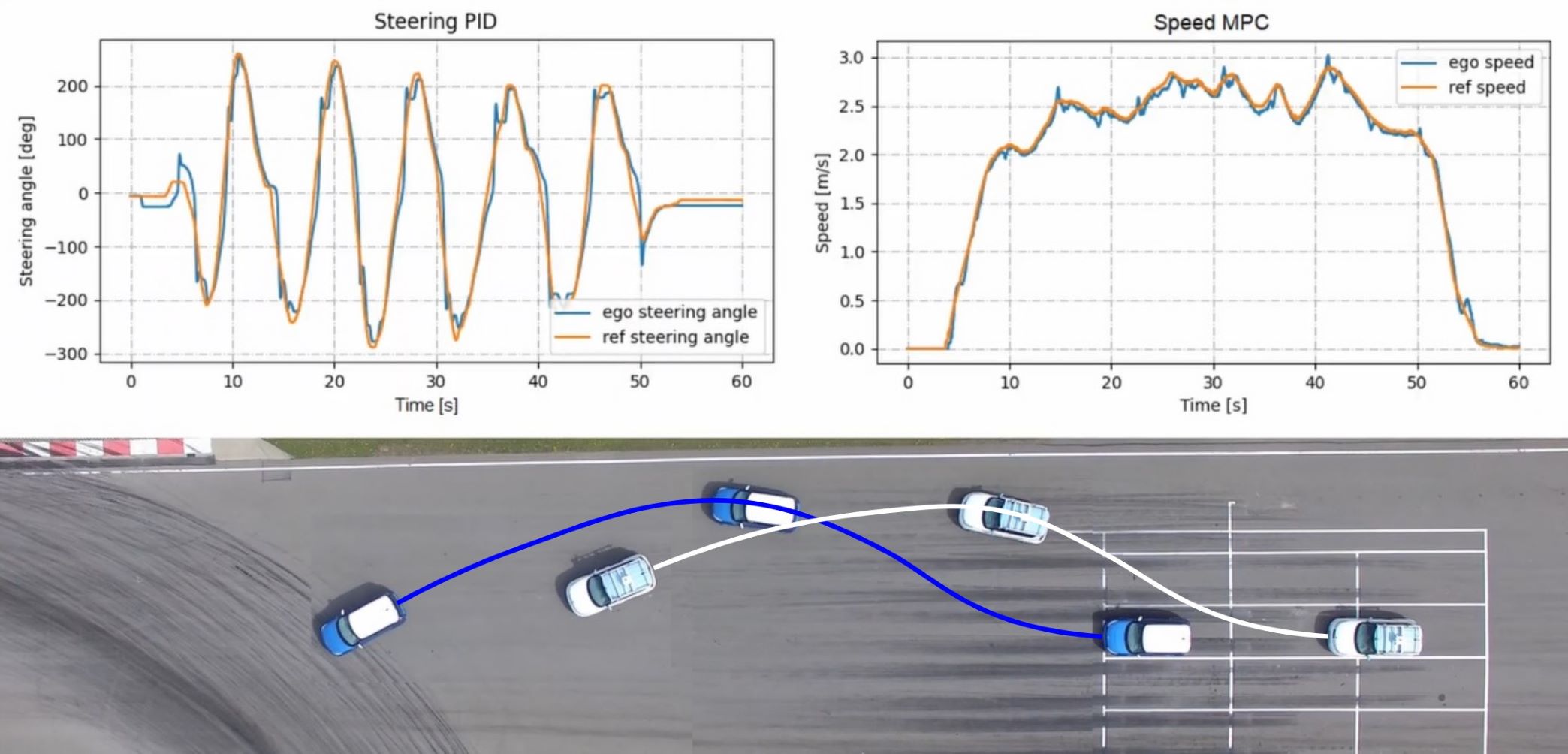}}
    \caption{Racetrack test showcasing synchronised driving via exchanged kinematics data between two vehicles, using an MPC for speed control and a PID controller for steering.}
    \label{fig:driving-test}
\end{figure*}

During the driving test, we align the two vehicles on the road and manually drive the \textit{Blue Kia}, allowing the \textit{White Kia} to autonomously replicate its driving pattern. Our MPC algorithm, described in \cref{sec:state-sync-mpc}, is used for longitudinal control, while a PID controller is used for steering torque for lateral control. We provide a graphical representation of the driving test in \cref{fig:driving-test}. The results indicate that the automated vehicle can accurately track the driving pattern of the reference vehicle, achieving a root-mean-square speed error of 0.06~m/s and a maximum error of 0.23~m/s.

%% file: sections/conclusions.tex
\section{Conclusions} \label{sec:conclusions}

This paper presents a versatile software and hardware platform for cooperative, connected, and autonomous mobility research. We provide a detailed overview of the hardware components and their interactions, along with insights into the custom drive-by-wire system, architecture, safety features, and additional modules used to implement the platform. Moreover, we emphasise elements that are less frequently discussed in similar publications. These include the interplay of virtual and real testing environments, wireless connectivity as part of the perception stack, and a driving environment that supports research of country border-crossing applications. Future compatibility and ease of development are facilitated through design considerations such as modularity, in both hardware and software, and the integration of ROS 2. We posit that our CCAM platform provides a convenient environment, both inside and outside of the vehicles for further research and development of context-aware autonomous vehicles.

A key area for future research is the application of 5G Standalone (SA) network capabilities. These technologies have the potential to significantly improve communication speed and reliability through the use of network slicing and multi-access edge computing (MEC). Our research group is working on implementing a vehicle string control system that enables dynamic participant joining and leaving through a network service designed for seamless service continuity across country borders. To achieve this, we plan to extend our model-predictive controller to provide robustness against packet loss by sharing predicted control values between participants. In addition, we are researching sensor-sharing solutions that enable our vehicles to react to distant events such as traffic jams and emergency services before they are perceived by in-vehicle sensors. While advancements in cellular communication are expected to improve the decision-making capabilities of autonomous vehicles, widespread adoption will necessitate active collaboration with mobile network operators to render these technologies accessible to the public.

%% file: bibliography.bib
@article{ismagilova2019smart,
  title={Smart cities: Advances in research—An information systems perspective},
  author={Ismagilova, Elvira and Hughes, Laurie and Dwivedi, Yogesh K and Raman, K Ravi},
  journal={International journal of information management},
  volume={47},
  pages={88--100},
  year={2019},
  publisher={Elsevier}
}

@article{certad2023jku,
  title={Jku-its automobile for research on autonomous vehicles},
  author={Certad, Novel and Morales-Alvarez, Walter and Novotny, Georg and Olaverri-Monreal, Cristina},
  journal={arXiv preprint arXiv:2301.06422},
  year={2023}
}

@article{mehr2022x,
  title={X-CAR: An Experimental Vehicle Platform for Connected Autonomy Research Powered by CARMA},
  author={Mehr, Goodarz and Ghorai, Prasenjit and Zhang, Ce and Nayak, Anshul and Patel, Darshit and Sivashangaran, Shathushan and Eskandarian, Azim},
  journal={arXiv preprint arXiv:2204.02559},
  year={2022}
}

@article{brown2022development,
  title={Development of an energy efficient and cost effective autonomous vehicle research platform},
  author={Brown, Nicholas E and Rojas, Johan F and Goberville, Nicholas A and Alzubi, Hamzeh and AlRousan, Qusay and Wang, Chieh and Huff, Shean and Rios-Torres, Jackeline and Ekti, Ali Riza and LaClair, Tim J and others},
  journal={Sensors},
  volume={22},
  number={16},
  pages={5999},
  year={2022},
  publisher={MDPI}
}

@article{camara2022openpodcar,
  title={OpenPodcar: an Open Source Vehicle for Self-Driving Car Research},
  author={Camara, Fanta and Waltham, Chris and Churchill, Grey and Fox, Charles},
  journal={arXiv preprint arXiv:2205.04454},
  year={2022}
}

@article{arango2020drive,
  title={Drive-by-wire development process based on ros for an autonomous electric vehicle},
  author={Arango, J Felipe and Bergasa, Luis M and Revenga, Pedro A and Barea, Rafael and L{\'o}pez-Guill{\'e}n, Elena and G{\'o}mez-Hu{\'e}lamo, Carlos and Araluce, Javier and Guti{\'e}rrez, Rodrigo},
  journal={Sensors},
  volume={20},
  number={21},
  pages={6121},
  year={2020},
  publisher={MDPI}
}

@inproceedings{novickis2020functional,
  title={Functional architecture for autonomous driving and its implementation},
  author={Novickis, Rihards and Levinskis, Aleksandrs and Kadikis, Roberts and Fescenko, Vitalijs and Ozols, Kaspars},
  booktitle={2020 17th Biennial Baltic Electronics Conference (BEC)},
  pages={1--6},
  year={2020},
  organization={IEEE}
}

@article{bunte2014central,
  title={Central vehicle dynamics control of the robotic research platform robomobil},
  author={B{\"u}nte, Tilman and Ho, Lok Man and Satzger, Clemens and Brembeck, Jonathan},
  journal={ATZelektronik worldwide},
  volume={9},
  number={3},
  pages={58--64},
  year={2014},
  publisher={Springer}
}

@inproceedings{gand2020lightweight,
  title={A Lightweight Virtualisation Platform for Cooperative, Connected and Automated Mobility.},
  author={Gand, Fabian and Fronza, Ilenia and El Ioini, Nabil and Barzegar, Hamid R and Van Thanh Le and Pahl, Claus},
  booktitle={VEHITS},
  pages={211--220},
  year={2020}
}

@article{kato2015open,
  title={An open approach to autonomous vehicles},
  author={Kato, Shinpei and Takeuchi, Eijiro and Ishiguro, Yoshio and Ninomiya, Yoshiki and Takeda, Kazuya and Hamada, Tsuyoshi},
  journal={IEEE Micro},
  volume={35},
  number={6},
  pages={60--68},
  year={2015},
  publisher={IEEE}
}

@inproceedings{scholliers2018development,
  title={Development of an automated vehicle as an innovation platform},
  author={Scholliers, Johan and Kutila, Matti and Virtanen, Ari and Pyyk{\"o}nen, Pasi},
  booktitle={25th ITS World Congress: Quality of life},
  pages={TP--1505},
  year={2018},
  organization={ERTICO-ITS Europe}
}

@inproceedings{avedisov2018experimental,
  title={Experimental verification platform for connected vehicle networks},
  author={Avedisov, Sergei S and Bansal, Gaurav and Kiss, Adam K and Orosz, G{\'a}bor},
  booktitle={2018 21st International Conference on Intelligent Transportation Systems (ITSC)},
  pages={818--823},
  year={2018},
  organization={IEEE}
}

@article{feher2020highly,
  title={Highly Automated Electric Vehicle Platform for Control Education},
  author={Feh{\'e}r, {\'A}rp{\'a}d and Aradi, Szil{\'a}rd and B{\'e}cs, Tam{\'a}s and G{\'a}sp{\'a}r, P{\'e}ter},
  journal={IFAC-PapersOnLine},
  volume={53},
  number={2},
  pages={17296--17301},
  year={2020},
  publisher={Elsevier}
}

@inproceedings{reke2020self,
  title={A self-driving car architecture in ROS2},
  author={Reke, Michael and Peter, Daniel and Schulte-Tigges, Joschua and Schiffer, Stefan and Ferrein, Alexander and Walter, Thomas and Matheis, Dominik},
  booktitle={2020 International SAUPEC/RobMech/PRASA Conference},
  pages={1--6},
  year={2020},
  organization={IEEE}
}

@inproceedings{kronauer2021latency,
  title={Latency analysis of ROS2 multi-node systems},
  author={Kronauer, Tobias and Pohlmann, Joshwa and Matth{\'e}, Maximilian and Smejkal, Till and Fettweis, Gerhard},
  booktitle={2021 IEEE International Conference on Multisensor Fusion and Integration for Intelligent Systems (MFI)},
  pages={1--7},
  year={2021},
  organization={IEEE}
}

@inproceedings{maruyama2016exploring,
  title={Exploring the performance of ROS2},
  author={Maruyama, Yuya and Kato, Shinpei and Azumi, Takuya},
  booktitle={Proceedings of the 13th International Conference on Embedded Software},
  pages={1--10},
  year={2016}
}

@inproceedings{munir2018autonomous,
  title={Autonomous vehicle: The architecture aspect of self driving car},
  author={Munir, Farzeen and Azam, Shoaib and Hussain, Muhammad Ishfaq and Sheri, Ahmed Muqeem and Jeon, Moongu},
  booktitle={Proceedings of the 2018 International Conference on Sensors, Signal and Image Processing},
  pages={1--5},
  year={2018}
}

@article{rassolkin2018development,
  title={Development case study of first Estonian Self-driving car ISEAUTO},
  author={Rass{\~o}lkin, Anton and Sell, Raivo and Leier, Mairo},
  journal={The Scientific Journal of Riga Technical University-Electrical, Control and Communication Engineering},
  volume={14},
  number={1},
  pages={81--88},
  year={2018}
}

@inproceedings{pardo2003omg,
  title={Omg data-distribution service: Architectural overview},
  author={Pardo-Castellote, Gerardo},
  booktitle={23rd International Conference on Distributed Computing Systems Workshops, 2003. Proceedings.},
  pages={200--206},
  year={2003},
  organization={IEEE}
}

@inproceedings{lee1996development,
  title={Development of autonomous test vehicle using image processing},
  author={Lee, Dong-Min and Kim, Dong-Ho and Kim, Byung-Soo and Moon, Soon-Hwan and Han, Min-Hong},
  booktitle={Proceedings of Conference on Intelligent Vehicles},
  pages={409--414},
  year={1996},
  organization={IEEE}
}

@article{dickmanns1987autonomous,
  title={Autonomous high speed road vehicle guidance by computer vision},
  author={Dickmanns, Ernst D and Zapp, Alfred},
  journal={IFAC Proceedings Volumes},
  volume={20},
  number={5},
  pages={221--226},
  year={1987},
  publisher={Elsevier}
}

@inproceedings{ulmer1994vita,
  title={Vita ii-active collision avoidance in real traffic},
  author={Ulmer, Berthold},
  booktitle={Proceedings of the Intelligent Vehicles' 94 Symposium},
  pages={1--6},
  year={1994},
  organization={IEEE}
}

@inproceedings{dickmanns1994seeing,
  title={The seeing passenger car'VaMoRs-P'},
  author={Dickmanns, Ernst Dieter and Behringer, Reinhold and Dickmanns, Dirk and Hildebrandt, Thomas and Maurer, Markus and Thomanek, Frank and Schiehlen, Joachim},
  booktitle={Proceedings of the Intelligent Vehicles' 94 Symposium},
  pages={68--73},
  year={1994},
  organization={IEEE}
}

@inproceedings{jochem1995pans,
  title={PANS: A portable navigation platform},
  author={Jochem, Todd and Pomerleau, Dean and Kumar, Bala and Armstrong, Jeremy},
  booktitle={Proceedings of the Intelligent Vehicles' 95. Symposium},
  pages={107--112},
  year={1995},
  organization={IEEE}
}

@online{5g-routes,
  title = {{5G-ROUTES}},
  year = 2020,
  url = {https://www.5g-routes.eu},
  urldate = {2023-03-13}
}

@online{5g-blueprint,
  title = {{5G-BLUEPRINT}},
  year = 2020,
  url = {https://www.5gblueprint.eu},
  urldate = {2023-03-13}
}

@online{5g-mobix,
  title = {{5G-MOBIX}},
  year = 2018,
  url = {https://www.5g-mobix.com},
  urldate = {2023-03-13}
}

@online{5g-carmen,
  title = {{5G-CARMEN}},
  year = 2018,
  url = {https://5gcarmen.eu},
  urldate = {2023-03-13}
}

@online{ai4csm,
  title = {{AI4CSM}},
  year = 2021,
  url = {https://ai4csm.eu},
  urldate = {2023-03-13}
}

@online{augmented-ccam,
  title = {{AUGMENTED CCAM}},
  year = 2022,
  url = {https://www.ccam.eu/projects/augmented-ccam},
  urldate = {2023-03-13}
}

@online{oscc-github,
  title = {{GitHub PolySync/OSCC}},
  year = 2017,
  url = {https://github.com/PolySync/OSCC},
  urldate = {2023-03-13}
}

@techreport{5g-routes-report,
  author = {Artūrs Lindenbergs and Kaspars Kalnins and Diana Kreivina and Inga Toleika and Armands Meirans and Helmut Zaglauer and Johan Schollers and Muhhamad Alam},
  title = {{D1.3 Report on 5G-ROUTES CAM deployment plan and longer-term deployment strategies}},
  institution = {5G-ROUTES},
  year = {2020},
  month = {12}
}

@online{asn1c,
  title = {{GitHub vlm/asn1c}},
  year = 2005,
  url = {https://github.com/vlm/asn1c},
  urldate = {2023-03-13}
}

@online{roadrunner,
  title = {{RoadRunner}},
  year = 2023,
  url = {https://www.mathworks.com/products/roadrunner.html},
  urldate = {2023-03-13}
}

@techreport{ETSI_CAM,
  title = "{ITS; Vehicular communications; Basic Set of Applications; Part 2: Specification of Cooperative Awareness Basic Service}",
  author = "{ETSI}",
  year = "2019",
  institution = "{European Telecommunications Standards Institute}",
  type = "{Technical Specification}",
  number = "{ETSI EN 302 637-2 V1.4.1}"
}
